\documentclass[preprint,9pt]{elsarticle}

\usepackage{amssymb,amsthm,mathtools,mathabx,mathrsfs} 
\usepackage[cal=boondoxo]{mathalfa} 
\usepackage{hyperref,url} 
\usepackage{makecell}
\usepackage{slashed}
\usepackage{stmaryrd}
\usepackage{comment}
\usepackage{fancyvrb}
\biboptions{numbers}
\usepackage{setspace}
\usepackage{xcolor,color} 
\usepackage{graphicx}
\usepackage{subcaption}
\usepackage[linesnumbered,ruled,vlined]{algorithm2e}

\usepackage{enumitem} 
\usepackage[normalem]{ulem}
\journal{Journal of Computational and Applied Mathematics}

\usepackage{bm}                 
\DeclareMathOperator*{\argmin}{\textrm{argmin}\,} 
\def\Diag            {\textrm{Diag}} 	 
\def\st              {\,\textrm{s.t.}\,} 
\def\tr              {\textrm{Tr}}       
\def\zeros           {\bm{0}}            
\def\IN {\mathbb{N}} 		
\def\IR  {\mathbb{R}} 	    
\def\IRm {\IR^{m}} 			
\def\IRn {\IR^{n}} 			
\def\IRr {\IR^{r}} 			
\def\IRmn{\IR^{m \times n}} 
\def\IRnn{\IR^{n \times n}} 
\def\IRmr{\IR^{m \times r}} 
\def\IRrn{\IR^{r \times n}} 

\def\A {\bm{A}}
\def\B {\bm{B}}

\def\E {\bm{E}}

\def\H {\bm{H}}

\def\M {\bm{M}}

\def\R {\bm{R}}

\def\W {\bm{W}}
\def\X {\bm{X}} 
\def\Y {\bm{Y}}

\def\b {\bm{b}}

\def\g {\bm{g}}
\def\h {\bm{h}}

\def\s {\bm{s}}

\def\u {\bm{u}}
\def\v {\bm{v}}
\def\w {\bm{w}}

\def\cL { \mathcal{L}}

\def\cN { \mathcal{N}}

\def\cP { \mathcal{P}}

\def\cR { \mathcal{R}}

\def\cU { \mathcal{U}}

\def\bPsi     {\boldsymbol{\Psi}}	
	
\def\blambda  {\boldsymbol{\lambda}}

\newtheorem{remark}{Remark}
\DeclareMathOperator*{\AlgoName}{\texttt{SHINBO}}  
\begin{document}
\begin{frontmatter}
\title{Sparse Hyperparametric Itakura-Saito Nonnegative Matrix Factorization via Bi-Level Optimization}

\author[inst1,inst2]{Laura Selicato}
\author[inst2]{Flavia Esposito\corref{mycorrespondingauthor}}
\cortext[mycorrespondingauthor]{Corresponding author}
\ead{flavia.esposito@uniba.it}
\author[inst3]{Andersen Ang}
\author[inst2]{Nicoletta Del Buono}
\author[inst4]{Rafa\l $ $ Zdunek}

\affiliation[inst1]{organization={Water Research Institute - National Research Council (IRSA-CNR)},
            country={Italy}}

\affiliation[inst2]{organization={Department of Mathematics, University of Bari Aldo Moro},
            country={Italy}}

\affiliation[inst3]{organization={School of Electronics and Computer Science, University of Southampton},
            country={UK}}

\affiliation[inst4]{organization={Faculty of Electronics, Photonics, and Microsystems, Wroc\l aw University of Science and Technology}, 
            country={Poland}
            }

\begin{abstract}
The selection of penalty hyperparameters is a critical aspect in Nonnegative Matrix Factorization (NMF), since these values control the trade-off between reconstruction accuracy and adherence to desired constraints. 
In this work, we focus on an NMF problem involving the Itakura-Saito (IS) divergence, which is particularly effective for extracting low spectral density components from spectrograms of mixed signals, and benefits from the introduction of sparsity constraints. 
We propose a new algorithm called SHINBO, which introduces a bi-level optimization framework to automatically and adaptively tune the row-dependent penalty hyperparameters, enhancing the ability of IS-NMF to isolate sparse, periodic signals in noisy environments.
Experimental results demonstrate that SHINBO achieves accurate spectral decompositions and demonstrates superior performance in both synthetic and real-world applications.
In the latter case, SHINBO is particularly useful for noninvasive vibration-based fault detection in rolling bearings, where the desired signal components often reside in high-frequency subbands but are obscured by stronger, spectrally broader noise.
By addressing the critical issue of hyperparameter selection, SHINBO improves the state-of-the-art in signal recovery for complex, noise-dominated environments. 
\end{abstract}

\begin{keyword}
Nonnegative Matrix Factorization,
Itakura-Saito divergence,
Hyperameter Optimization,
Sparsity,
Bi-level Optimization,
Dynamical System
\end{keyword}
\end{frontmatter}

\section{Introduction}
Nonnegative Matrix Factorization (NMF) is a dimensionality reduction technique that approximates a nonnegative data matrix as the product of two (lower-dimensional) nonnegative matrices. 
A key challenge in NMF is to determine penalty coefficients when additional constraints, such as sparsity or smoothness, are imposed \cite{gillis2020nonnegative}. 
These coefficients control the trade-off between reconstruction accuracy and constraint adherence, but their optimal values are highly dependent on both the dataset and the specifict application, making the selection process non-trivial. For example, \cite{corsetti2020nonnegative} proposes a variant of NMF that incorporates data-dependent penalties and introduces auxiliary constraints to enhance performance in tasks such as face recognition. Additionally, \cite{lanteri2011nonnegative} presents multiplicative algorithms for NMF that enforce non-negativity and flow preservation constraints while introducing regularizations to ensure smoothness or sparsity. Finally, \cite{fabregat2019solving} adapts a minimization scheme for functions with non-differentiable constraints, known as PALM, to solve NMF problems, yielding solutions that can be both smooth and sparse—two highly desirable properties. In this work, we rely on previous studies \cite{del2023bi,del2025penalty} in which the penalized problem is reformulated in a general form, and a strategy is proposed to automatically adjust the penalty coefficient.

Formally, let $\X \in \IRmn_+$ with $m,n \in \IN$ be a data matrix, NMF aims to approximate it as the product of $\W\in\IRmr_+$ and $\H\in\IRrn_+$ with $r \leq \min\{m,n\}$, so that $\X\approx \W\H$. 
In our problem, we want to solve
\begin{equation}\label{min_prob}
(\W^*,\H^*)
~\in
    \displaystyle\argmin_{\substack{\W \geq \zeros, \H \geq \zeros  }} 
    D_{\beta}(\X, \W\H) 
    + \mathcal{P}\big(
    \Diag(\blambda)\H
    \big) 
\end{equation}
with the objective function $D_\beta(\cdot,\cdot)$ being the $\beta$-divergence \cite{fevotte2011algorithms,kompass2007generalized}, assessing the quality of the reconstruction $\W\H$ in fitting $\X$
\footnote{We remind that the $\beta$-divergence for matrices can be computed element-wise as $D_\beta(\A,\B)=\sum_i{\sum_j{d_\beta(a_{ij},b_{ij})}}$, where the function $d_\beta$ for each $x,y \in \IR$ is defined as
\[
d_{\beta}(x,y)= \left \{
\begin{array}{lc}
\frac{1}{\beta(\beta-1)}(x^{\beta}+(\beta-1) y^{\beta}-\beta xy^{\beta-1}) &\beta\in \IR \setminus \{0,1\};
\vspace{2mm}
\\
x\log(\frac{x}{y})-x+y &\beta=1;
\vspace{2mm}
\\
\frac{x}{y}-\log(\frac{x}{y})-1 &\beta=0.
\end{array}
\right.
\]}.
The function $\cP : \IRrn \to \IR$ is a penalty term on $\H$ that enforces a particular constraint.
The vector $\blambda \in \IRr_+$ in \eqref{min_prob} contains nonnegative penalty hyperparameters $[\lambda_1, \dots, \lambda_r]$ associated to each row of $\H$.

In this study, we focus on a special case of the $\beta$-divergence within NMF. The $\beta$-divergence offers a tunable objective function that adapts to the statistics of the data: by varying $\beta$ it interpolates between the Euclidean distance ($\beta=2$), the Kullback-Leibler divergence ($\beta=1$), and—in the limit $\beta\to0$—the Itakura-Saito (IS) divergence. The IS divergence is scale-invariant and normalizes the reconstruction error by the signal magnitude, so weak spectral components (i.e. frequency components with low spectral density) contribute proportionally to the objective rather than being dominated by high-energy regions. As a results, IS-NMF is well suited to preserving fine, low-energy structures (e.g., weak higher-order harmonics, background noise, or faint events) that are often missed by Euclidean/KL objectives. Moreover, IS connects naturally to a multiplicative Gamma-noise model for power spectra, which aligns with the heavy-tailed, non-Gaussian behavior commonly encountered in audio and vibration spectrograms. 

Our target application is non-invasive, vibration-based fault detection in rolling bearings~\cite{wodecki2019novel,wodecki2019impulsive,wodecki2020separation,YANG2025118742,yang2025integrated}. The desired signal of interest (SOI) is periodic and impulsive, yet it typically resides in a high-frequency subband and is masked by much stronger, spectrally broad disturbances. We validate our approach on laboratory test-rig measurements of faulty bearings~\cite{GABOR2023110430}. Although time-domain perturbations are often modeled as additive i.i.d. Gaussian noise, after time-frequency transformation, the IS divergence provides a statistically well-matched criterion for factorizing spectrograms with multiplicative fluctuations.

Related work has applied $\beta$-NMF to source separation in audio~\cite{leplat2020blind} and established identifiability via \emph{minimum-volume} (MinVol) regularization. The “sufficiently scattered” condition~\cite{fu2018identifiability} relaxes separability assumptions, and subsequent studies showed that minimizing the convex-hull volume of basis columns can recover factors even in rank-deficient or nonseparable regimes~\cite{leplat2020blind}, thereby linking MinVol-NMF to practical uniqueness guarantees.

Methodologically, we apply NMF to the spectrogram of the measured signal. Let $\H$ denote the temporal activations. We expect the SOI to be captured by one row of $\H$ that exhibits sparse, periodic, spike-like activity, whereas rows associated with broadband disturbances are comparatively dense. To bias the factorization toward the SOI, we impose a row-sparsity-promoting penalty on $\H$ via the term $\cP(\Diag(\blambda)\H)$ in (\ref{min_prob}). We further adopt a bi-level scheme that adapts the hyperparameters $\blambda$ from the data, aligning the penalty strength with the latent structure of the estimated components.

\paragraph{\textbf{Contribution}}
The contribution of this work is twofold.
\begin{enumerate}
\item \textbf{New model}.
In this work, we present a new model for minimizing the Itakura-Saito divergence (Problem \eqref{min_prob}  with $\beta=0$) while penalizing rows of $\H$.
In particular
\begin{itemize}[leftmargin=10pt]
    \item The penalty hyperparameter is not known in advance.
    We treat this parameter as an optimization variable, embedding it in a bi-level optimization framework and solving it by a bi-level optimization method.

    \item The penalty hyperparameter is row-dependent.
    Note that Problem~\eqref{min_prob} is different from standard penalized NMF \cite{gillis2020nonnegative}, which applies the same penalty coefficient to all rows in the matrix $\H$.
    In Problem~\eqref{min_prob}, each row of $\H$ is penalized by its own penalty parameter.
\end{itemize}

\item \textbf{New algorithm}.
For Problem~\eqref{min_prob}, we present a new multiplicative update (see Equation~\eqref{H_up_fro_J_row}), and a way to automatically tune the penalty hyperparameter based on a bi-level strategy (see details in Section~\ref{sec:bilevel_general}). 
\end{enumerate}

\paragraph{\textbf{Paper Organization}}
We introduce the problem and overall algorithm framework in Section~\ref{sec:overall}.
In Section~\ref{sec:bilevel_general} we first review the bi-level optimization, then we discuss the details of the bi-level approach proposed in this work for solving the Problem~\ref{min_prob} in Section \ref{sec:algo}.
Experimental results on synthetic and real datasets are presented in Section~\ref{sec:exp}.
We conclude the paper in Section~\ref{sec:conl}, giving an outline of possible future directions.

\paragraph{\textbf{Notation}}
The symbol $\zeros$ denotes the zero matrix, the symbol $\E_{a\times b}$ is the all-in matrix size $a\times b$ with $a,b\in\IN$.
The notation $\v$ denotes a column vector and the notation $\underline{\v}$ means that $\v$ is a row vector.
On matrix, $\A^\top$ is the transpose of $\A$, and $\A^2 = \A\A$.
The symbol $\A \odot \B$ refers to the Hadamard (element-wise) product between $\A$ and $\B$ of conformal dimensions, and the symbol $\A \oslash \B$ with $\B \neq \zeros$ refers to the Hadamard division, and
we denote $\A^{\odot k}$ as the Hadamard power-$k$ of $\A$.
Given $n \in \IN$, we denote $[n] \coloneqq \{1,2,\ldots,n\}$.
The symbols $k,t \in \IN$ indicate iteration counters.
The symbol $\A^{k}$ refers to the variable $\A$ at the iteration $k$, $A_{ij}$ or $a_{ij}$ is the $(i,j)$th element of $\A$.
Lastly $\A_{i:}$ and $\A_{:j}$ are the $i$th and $j$th row and column of $\A$, respectively.

\section{The overall optimization framework of $\AlgoName$}\label{sec:overall}
In this section, we discuss the focus of the paper, Problem~\eqref{min_prob} and the overall framework of the proposed algorithm.

\paragraph{\textbf{The optimization problem}}
We focus on Problem~\eqref{min_prob} with $\beta = 0$. As $\cP$ we chose a particular penalty function, effective in increasing sparsity, that is the diversity measure $J$ \cite{cotter2005sparse,del2025penalty}. 
In a particular case, if the matrix is nonnegative, the diversity measure can be written as
\begin{equation}\label{J_row}
J(\A) = \sum_{i=1}^n \|\A_{i:}\|_1^2 = Tr(\A \E \A^\top),
\end{equation}
where $\mathbf{E}$ is the matrix of 1s. 
By properties of trace, Problem~\eqref{min_prob} becomes
\begin{equation}\label{probl}
(\W^*,\H^*,\blambda^*)
~\in
\argmin_{\substack{\W \geq \zeros, \H \geq \zeros \\ \blambda \geq \zeros}} 
D_{0}(\X, \W\H) 
+ \tr\big(\Diag(\blambda)^2 \H\E \H^\top\big)
.
\end{equation}
We remark that \eqref{probl} is a nonconvex minimization problem, in which finding global minima is NP-hard, therefore we are interested in finding local minima for the triple $(\W^*,\H^*,\blambda^*)$.

\paragraph{\textbf{The proposed optimization algorithm $\AlgoName$}}
We propose an algorithm, called $\AlgoName$, to find a local minima for Problem~\eqref{probl} as follows.

\begin{algorithm}[ht]
\SetAlgoLined
\textbf{Input:}{ $\X \in \IRmn_+$ and factorization rank $r$.}\\
Initialize $\W^{0} \in \IRmr_+$, $\H^{0} \in \IRrn_+$ and $\blambda^{0} \in \IRr_+$\\
\For{$k = 1,2,...$ }{
$\W^k = \text{update}(\X,\W^{k-1},\H^{k-1})$ 
\hfill \% classic MU-update
\\
\For{$l \in [r]$}{
$\underline{\h}_l^{k-1,0} = \underline{\h}_l^{k-1}, \lambda^{k-1,0}_l = \lambda^{k-1}_l$
\hfill \% initialization
\\
\For{$t \in [T]$}{
$\underline{\h}_l^{k,t} = \text{update} (\underline{\h}_l^{k,t-1},\W^k,\X,\lambda_l^{k})$ as in $\text{\eqref{H_up_fro_J_row}}$\\
$\frac{\partial \cR}{\partial \lambda_l}$ as in \eqref{hyppp}
\hfill \% hypergradient
}
$\blambda^{k,T} = \text{update}(\blambda^{k,T},\nabla_{\blambda}\cR(\blambda))$
\hfill \% projected gradient update
}
}
\textbf{Return }{ $\W,\H,\blambda$ at the last iteration}.
\caption{$\AlgoName$}
\label{alg1}
\end{algorithm}

Note that $\AlgoName$ is composed by two main parts: one devoted to the update of $\W$ (reviewed in the following) and the other one based on bi-level strategy to optimize simultaneously on $\H$ and $\blambda$.

\paragraph{\textbf{Update on W}}
The update of $\W$ can be done simply by the following multiplicative update \cite{fevotte2009nonnegative} as $\W = \W \odot  (\X\H^\top) \oslash  (\W\H\H^\top)$.

The following section introduces the bi-level update on $\H$ and $\blambda$ by a bi-level method, which is the main contribution of the paper.

%
%
\section{Bi-level Optimization for the subproblem}\label{sec:bilevel_general}
In this section, we discuss the steps for updating $\H$ and $\blambda$ in Algorithm~\ref{alg1}.
Given a fixed $\W^k$, we have the following optimization subproblem
\begin{equation}\label{prob_HL}
(\H^*,\blambda^*)
~\in
    \argmin_{\substack{\H \geq \zeros, \blambda \geq \zeros}} 
    D_{0}(\X, \W\H) 
    + \tr\big(\Diag(\blambda)^2 \H\E \H^\top\big)
    .
\end{equation}
The core idea of this paper is to solve Problem~\eqref{prob_HL} as a bi-level problem, in which we incorporate the problem of tuning hyperparameter $\blambda$ simultaneously into the update of $\H$. 
To do so, first we review the general theory of bi-level optimization
and its application to Problem \eqref{prob_HL}. 

\textbf{\textit{Section organization and overview of the approach.}}
Under a fixed and given $\W^k$, the goal is to obtain an updated version of $\H$ and $\blambda$ that approximately solves Problem~\eqref{prob_HL}.
The bi-level approach has the following steps:
\begin{enumerate}[leftmargin=15pt]
    \item We first rewrite the constrained optimization problem \eqref{prob_HL} by a bi-level optimization problem for $(\underline{\h}_l,\lambda_l)$, with $l \in [r]$, see \eqref{subproblem:bilevel}.
    \item 
    The inner problem (IP) in Problem~\eqref{subproblem:bilevel}  
    is then approximated by the solution of a dynamical system.
    See Problem~\eqref{IVP}.
    \item We then solve Problem~\eqref{IVP} to obtain a solution for $\underline{\h}_l$, and also the hypergradient at the last time point $T$.
    See \eqref{hypergradient}.
    \item Lastly, we use the hypergradient to obtain the solution $\blambda$ by a gradient descent approach.
    See \eqref{min2}.
\end{enumerate}
We now proceed to discuss each of the steps below.

\paragraph{\textbf{1. Bi-level formulation}}
In Algorithm~\ref{alg1}, the update of $\H$ and $\blambda$ is performed in $r$ steps, where each step is aimed to update the $l$th component $(\underline{\h}_l, \lambda_l)$.
This is achieved by solving the following bi-level problem 
\begin{equation}\label{subproblem:bilevel}
\begin{array}{rcl}\hspace{-5mm}
\displaystyle
\min_{\lambda_l \geq 0} 
\left\{\hspace{-1.5mm}
    \begin{array}{rl}
    \displaystyle
    \mathcal{r}(\lambda_l) 
    =& \hspace{-3mm}
    \displaystyle \inf_{\underline{\h}_l(\lambda_l) } 
                  D_2\big(\X,\R+\w_l\underline{\h}_l(\lambda_l)\big) \qquad \text{(OP)}
                  \vspace{1mm}
                  \\
    &\hspace{-3mm} \st \,\, \displaystyle \underline{\h}_l(\lambda_l) \in \argmin_{\u \in \IRn_+} D_0(\X,\R+\w_l\underline{\u})+\lambda_l^2\|\underline{\u}\|_1^2 \quad \text{(IP)}
    \end{array}
    \hspace{-2mm}
\right\},
\end{array}
\end{equation}
where the matrix $\R$ is the residual obtained isolating in $\W\H$ the $l$th component of $\w_l\in\IRm$ (column of $\W$) and $\underline{\h}_l \in \IRn$ (row of $\H$), which is
\[
\R = \X - \sum_{j \neq \ell} \w_j \underline{\h}_j.
\]
The function $\mathcal{r}: \IR \to \IR$ is called the Response function of the outer problem related to $\underline{\h}_l$. 
In the outer problem, the objective $D_2$ is the $\beta$-divergence with $\beta = 2$, which is the Frobenius norm.
The inner problem is represented by the $\beta$-divergence with $\beta = 0$, which is the Itakura-Saito divergence $D_0$, regularized by the row-wise diversity measure defined in \eqref{J_row}. 

\begin{remark}
Note that the squared $\ell_1$-norm in the inner problem on $\u$ can be seen as a non-smooth regularization term and therefore proximal gradient descent can be applied on $\u$ (see \cite{evgeniou2015regularized} and \cite[Lemma 6.70]{beck2017first}), however, their approach is applied to convex objective function, where here in \eqref{subproblem:bilevel} the objective function (the IS-divergence) is possibly nonconvex \cite{fevotte2009nonnegative} thus proximal gradient descent do not have convergence guarantee.    
\end{remark}

\paragraph{\textbf{2. Dynamical system approach on H}}
An approach to solve the bi-level Problem \eqref{subproblem:bilevel} over $\H$ is to replace the inner problem with a dynamical system~\cite{franceschi2017forward,Mc} and compute an approximation solution.\\
We now omit the $k$ index in $\underline{\h}_l^{k,t}$, due to the fact that the update focuses on the iteration over $t$ under a constant $k$.\\
Given $\underline{\h}_l^{0}$ (which depends implicitly on $\lambda_l$), we build a dynamical system (IVP-$\Phi$) in the form of a discrete initial value problem as
\begin{equation}\label{IVP}
\begin{cases}
\underline{\h}_l^t= \Phi_t(\underline{\h}_l^{t-1}, \lambda_l), \quad t \in [T] 
\\
\underline{\h}_l^{0} = \Phi_0(\lambda_l)
\end{cases}
\tag{IVP-$\Phi$}
\end{equation}
where $\Phi_t : \IRn \times \IR \rightarrow \IRr$ is a smooth map for $ t \in [T]$.

Bi-level optimization uses the IVP-$\Phi$ to approximate the solution of the Problem \eqref{subproblem:bilevel}.
We do so by solving the following minimization problem 
\begin{equation}\label{min2}
\begin{array}{rl}
\displaystyle 
\lambda_l^* 
=
\argmin_{\lambda_l} & \mathcal{r}(\lambda_l)
\\
\st ~~ &
\underline{\h}_l^{t} = \Phi_t(\underline{\h}_l^{t-1}, \lambda_l) \quad \text{for } t \in [T],
\end{array}
\end{equation}
in which we approximate the solution of the inner problem with the solution of the dynamical system. 
This is possible because problem~\eqref{min2} satisfies the existence and convergence theorems as proved in \cite{del2023bi}. 

As a preview, we will derive $\Phi_t(\underline{\h}_l^{t-1}, \lambda_l)$ in Section~\ref{sec:algo}.

\paragraph{\textbf{3. Hypergradient}}
To find the solution $\blambda^*$ for Problem~\eqref{prob_HL}, we solve the problem formed by joining all the Response functions $\mathcal{r}(\lambda_l)$ in \eqref{subproblem:bilevel} as
\begin{equation}\label{probl_big_R}
    \argmin_{\blambda \in \IRr_+} 
    \bigg\{ \cR(\blambda) \coloneqq  \sum_j \mathcal{r} (\lambda_l) \bigg\}.
\end{equation}
We would like to compute the hypergradient, i.e., the gradient of $\cR$ with respect to $\blambda$, in order to use a gradient descent approach on $\blambda$.
The hypergradient $\nabla_{\blambda}\cR$, by using chain rule, is
\begin{equation}\label{hypergradient}
\frac{\partial \cR}{\partial \lambda_l} 
= 
\dfrac{\partial \mathcal{r}}{\partial \lambda_l} +
\frac{\partial \mathcal{r}}{\partial \underline{\h}_l^{T}} \cdot \frac{d \underline{\h}_l^{T}}{d \lambda_l}, 
\quad l \in [r].
\tag{hypergrad}
\end{equation}
Note that $\underline{\h}_l^{T}$ denotes the row-vector $\underline{\h}_l$ at the time $T$.

It is well known that the computation of the hypergradient can be done using Reverse-Mode Differentiation (RMD) or Forward-Mode Differentiation (FMD).
Since RMD requires storing specific variables across all iterations and indices in memory, in this work, we use FMD, making it more suitable for scenarios where the total quantity of interest is small. For details we refer the reader to \cite{del2023bi,franceschi2017forward}.

\textbf{Forward-Mode}.
FMD computes the differentiation in \eqref{hypergradient} using the chain rule.
Function $\Phi_t$ for $t \in [T]$ depends on $\lambda_l$ explicit and on $\underline{\h}_l^{t-1}$ implicitly, then we have the derivative
\[
\frac{d \underline{\h}_l^{t}}{d \lambda_l}
= 
\frac{\partial \Phi_t(\underline{\h}_l^{t-1},\lambda_l)}{\partial \underline{\h}_l^{t-1}}
\cdot
\frac{d \underline{\h}_l^{t-1}}{d \lambda_l}+\frac{\partial \Phi_t(\underline{\h}_l^{t-1}, \lambda_l)}{\partial \lambda_l}.
\]
Let $\s^t = \dfrac{d \underline{\h}_l^{t}}{d \lambda_l}$, then each FMD iterate behaves as
\begin{equation}\label{eqdiff_As}
\begin{cases}
    \s^t = \A_t \s^{t-1}+ \b_t, \quad t \in [T]
    \\
    \s^0 = \b_0
\end{cases}
\end{equation}
where $\A_t  = \dfrac{\partial{\Phi_t(\underline{\h}_l^{t-1}, \lambda_l)}}{\partial{\underline{\h}_l^{t-1}}} \in \IRnn $ and 
$
\b_t =  \dfrac{\partial{\Phi_t(\underline{\h}_l^{t-1}, \lambda_l)}}{\partial{\lambda_l}}
\in \IRn$.
Now \eqref{hypergradient} can be expressed as
\begin{equation}\label{Hypergradiente_R}
\dfrac{\partial \cR}{\partial \lambda_l} 
= \langle \underline{\g}^T, \s^T \rangle \in \IR
\quad
\text{where} 
\quad
\underline{\g}^T
= \dfrac{\partial \mathcal{r}}{\partial \underline{\h}_l } \in \IRn.
\end{equation}
We note that $\underline{\g}^T$ is 
denoting the row-vector $\underline{\g}$ at the time $T$.
Lastly, the solution of Problem \eqref{eqdiff_As} solves the following equation
\begin{equation}\label{hyppp}
\dfrac{\partial \cR}{\partial \lambda_l} 
=
\dfrac{\partial \mathcal{r}}{\partial \underline{\h}_l^{T}} \left(
\b_T + \sum_{t=0}^{T-1}\bigg(\prod_{s=t+1}^T \A_s \bigg) \b_t
\right).
\end{equation}

\paragraph{\textbf{4. Update of $\blambda$}}
Given all the components $\underline{\h}_1^T,\underline{\h}_2^T,\ldots,\underline{\h}_r^T$ at the last time point of the dynamic system and the hypergradient (discussed previously), we update $\blambda$ with a projected gradient descent approach 
as
$
\blambda = \big[\blambda - \alpha \nabla_{\blambda}\cR(\blambda)\big]_+,
$
for a pre-defined stepsize $\alpha > 0$.


%
%
%
\section{Derivation of the algorithm SHINBO}\label{sec:algo}
We now introduce the overall bi-level optimization approach, discussed in the previous section, to the Subproblem~\eqref{prob_HL}.
We use the method of partial Lagrangian multiplier, which is applied only on $\H$.  
First, let $\bPsi \in \IRrn_+$ be the matrix of Lagrangian multipliers associated to the nonnegative constraints of $\H$, then the expression of the (partial) Lagrangian of Subproblem~\eqref{prob_HL} is
\[
\cL(\H)
= 
D_0(\X,\W\H)
+ \tr\big(\Diag(\blambda)^2\H\E\H^\top\big)+\tr\big(\bPsi\H^\top\big).
\]
Recall that $\M^{\odot2}$ denotes the Hadarmard power of $\M$, then the partial derivative of $\cL$ with respect to $\H$ is
\[
\dfrac{\partial{\mathcal{L}}}{\partial \H} 
= 
\W^\top\Big(
(\W\H)^{\odot -2}(\W\H-\X)
\Big)
+
2\Diag(\blambda)^2\H \E+\bPsi.
\]
Now denote $H_{ij}$ the $(i,j)$th element of the matrix $\H$, and recall that $\odot$ is the Hadarmard product, then by the complementary slackness $\Psi_{ij} H_{ij} = 0$ in the KKT conditions, we get 
\[
    \Big[
    \W^\top\Big(
    (\W\H)^{\odot -2} \odot (\W\H-\X)
    \Big) 
    \Big]_{ij}
    H_{ij}
+
    2
    \Big[
    \Diag(\blambda)^2\H\E
    \Big]_{ij} H_{ij} 
+ 
\Psi_{ij} H_{ij}
= 
0.
\]
These equations lead to the multiplicative update
\begin{equation}\label{H_up_fro_J}
H_{ij} 
= 
H_{ij}
\Big[
\W^\top(\W\H)^{\odot -2}\X
\Big]_{ij}
\bigg\slash
\Big[
\W^\top(\W\H)^{\odot -1}+2\Diag(\blambda)^2\H\E
\Big]_{ij}
,
\end{equation}
where the division is performed element-wise as the Hadamard division $\oslash$.
By fixing $l \in \{1, \dots, r\}$, the update in row-wise format for the $l$th row of $\H$ can be rewritten equivalently as
\begin{equation}\label{H_up_fro_J_row}
\underline{\h}^t_l
= 
\underline{\h}^{t-1}_l
\odot 
\dfrac{
    \Big(
    \W^\top(\W\H)^{\odot -2}\X
    \Big)_{l:}^{t-1}
}
{
    \Big(
    \W^\top \odot (\W\H)^{\odot -1} 
    \Big)_{l:}
    +
    2(\lambda_l^{t-1})^2\|\underline{\h}^{t-1}_l\|_1 \E_{l:}
}.
\end{equation}
We remark that Equation~\eqref{H_up_fro_J_row} is one of the main contribution of this paper.

\begin{remark}\label{remark:why_partial_Lag}
The key idea in this methodology is that $\Phi_t(\underline{\h}_l^{t-1}, \lambda_l) $ in the dynamical system \eqref{IVP} is the right hand side of the update~\eqref{H_up_fro_J_row}.
\end{remark}

\subsection{The implementation of the bi-level approach in SHINBO}


Following the procedure and the discussion in Section~\ref{sec:bilevel_general}, we consider $\Phi_t(\underline{\h}_l^{t-1}, \lambda_l) $ in the dynamical system \eqref{IVP} (represented by the multiplicative update~\eqref{H_up_fro_J_row}) and we compute $\A^t= \dfrac{\partial{\Phi_t}}{\partial{\underline{\h}^{t-1}_l}} $ and $\b^t = \dfrac{\partial{\Phi_t}}{\partial{\lambda_l}}$ for $t \in [T]$ required for the FMD.

\begin{itemize}[leftmargin=15pt]
    \item The computation of $\A^t= \dfrac{\partial{\Phi_t}}{\partial{\underline{\h}^{t-1}_l}}$ gives a diagonal matrix
    \[
    A_{jj}^t = 
    \dfrac{N_{lj}}{D_{lj}}-h_{lj}
    \left(
    \frac{\displaystyle2\sum_i^n{w^2_{il} \dfrac{x_{ij}}{(\W\H)_{ij}^3} D_{lj}-N_{lj}\sum_{i}^n}{
    \dfrac{w^2_{il}}{(\W\H)^2_{lj}}
    }+2\lambda_l^2N_{lj}}{D_{lj}^2}
    \right),
    \]
    where the derivative is computed with respect to the $(l,j)$th element of $\H$, and 
    \[
    N_{lj} = \big(\W^\top(\W\H)^{\odot -2}\X\big)_{lj},
    ~~
    D_{lj} = \big(\W^\top(\W\H)^{\odot -1}+2\Diag(\blambda)^2\H\E\big)_{lj}.
    \]

    \item The vector $\b_t$ is given by $\dfrac{\partial{\Phi_t}}{\partial{\lambda_l}} = - 4\lambda_l h^2_{lj} \dfrac{N_{lj}}{D_{lj}^2}
    $.
\end{itemize}

Finally, having chosen as an outer problem the Frobenius norm, $\underline{\g}^T$ in \eqref{Hypergradiente_R} for computing \eqref{hypergradient} is $\underline{\g}^T = -2\w_l^\top(\X-\R-\w_l\h_l)$.

\paragraph{\textbf{Computational cost}}
The computational cost of the proposed algorithm depends on the matrix-vector multiplications, as is usual in NMF algorithms, and on the bi-level framework, which requires the use of internal loops. The total computational cost of SHINBO is $\mathcal{O}(KTmnr)$, where $K$ is the number of iterations, and $T$ is the number of internal iterations. Although the proposed method appears to suffer from a high computational cost in construction, the gain in efficiency and accuracy in hyperparameter estimation justifies the high cost. Indeed, solving the NMF problem by running a well-known grid or random search methods requires more effort than SHINBO, which automatically selects the best hyperparameter vector for each row of $\H$.

%
%
\section{Experimental Results}\label{sec:exp}
In this section, we present the numerical results of comparing the proposed algorithm SHINBO with the multiplicative update (MU) algorithm in \cite{fevotte2009nonnegative}.
We also compare SHINBO with the penalized MU, which is the update \eqref{H_up_fro_J_row} under a fixed penalty hyperparameter $\blambda$ with $\lambda_1=\lambda_2=\ldots=\lambda_r$ for every row of $\H$.
We summarize their difference in the table below.

\begin{table}[h!]
\centering
\caption{The method compared in the experiment.}
\begin{tabular}{l|ll}
Algorithm & $\lambda$  &  $\lambda$ setting \\ \hline
MU \cite{fevotte2009nonnegative}      & 0   &constant
\\
Penalized MU  & 0.1  &constant for each column
\\
Penalized MU  & 0.5  &constant for each column
\\
SHINBO  & not required  & automatically optimized,\\
  & & different penalization for each column
\end{tabular}
\end{table}

\begin{figure}[h!]
    \centering
    \includegraphics[width=0.55\textwidth]{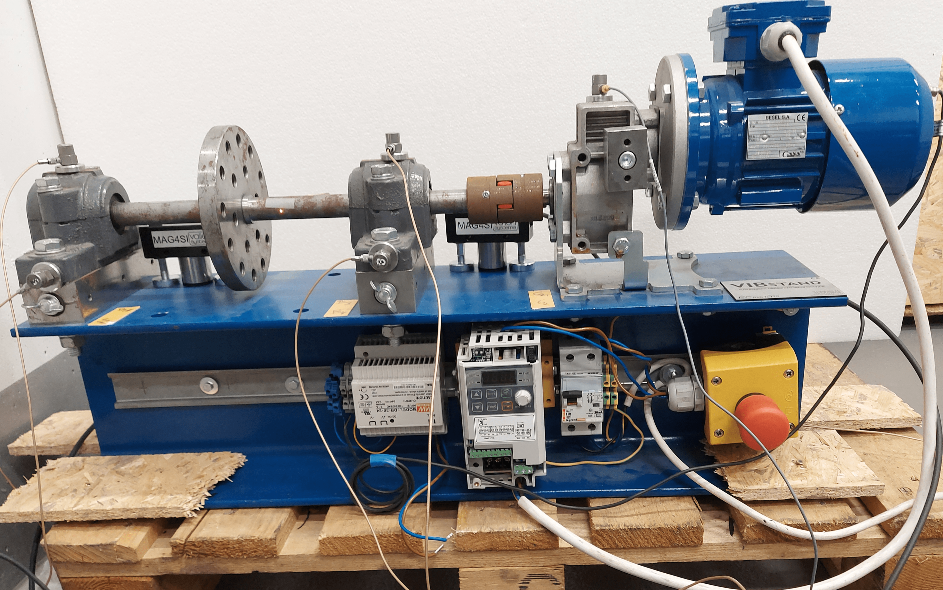}
    \caption{The test rig used in the experiment \cite{GABOR2023110430}.}
    \label{fig:rig}
\end{figure}

\begin{figure}[h!]
     \centering
        \includegraphics[width=0.88\textwidth]{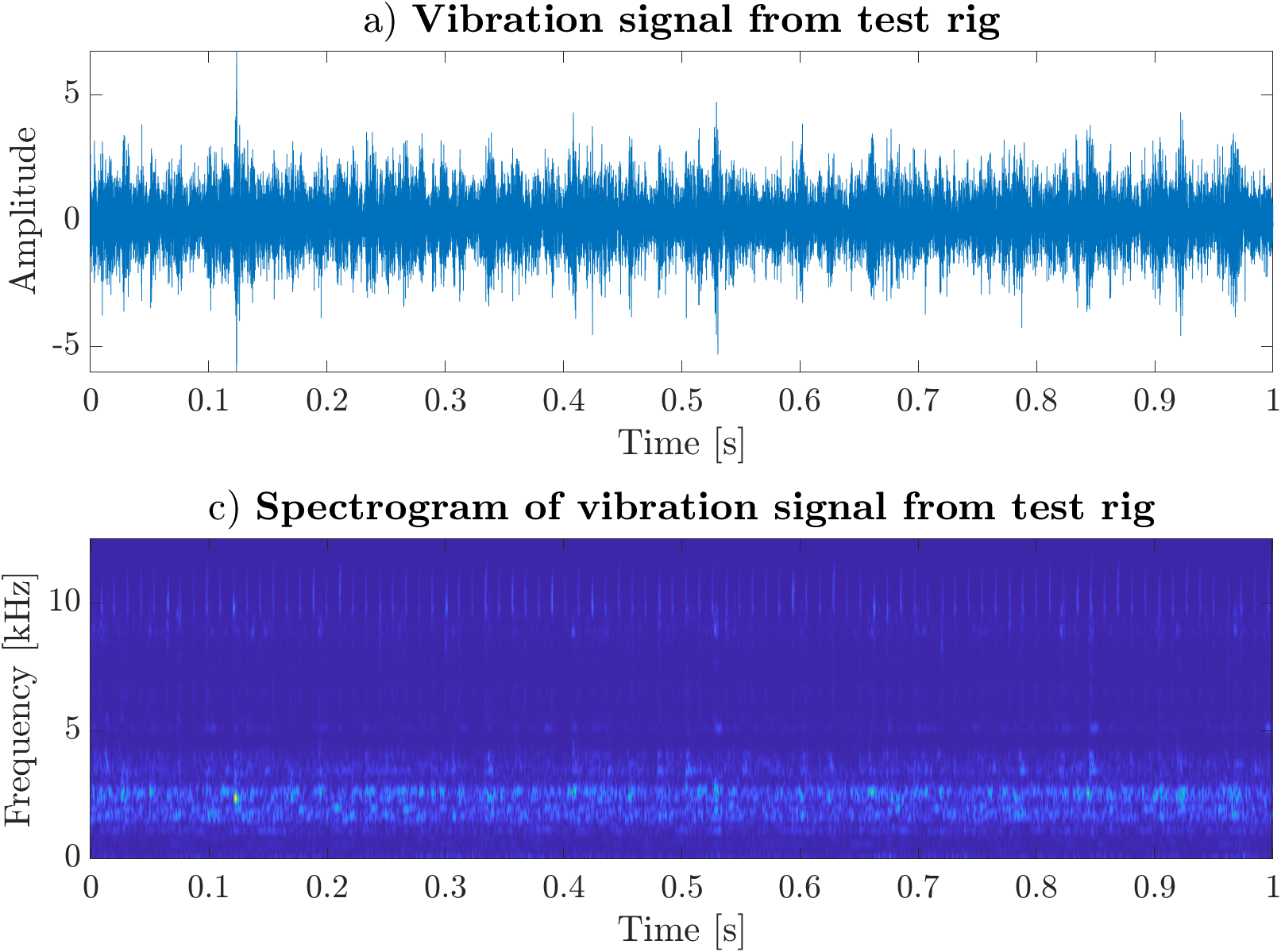}
     \caption{Recorded vibration signal and its spectrogram.}
     \label{fig:real_signals}
\end{figure}

\paragraph{\textbf{Datasets}}
We evaluate the algorithms on two datasets: one synthetically generated and the other obtained from real vibrational signals measured under laboratory conditions from damaged rolling bearing elements. 
\begin{itemize}[leftmargin=15pt]
\item The synthetic dataset is generated starting from a full-rank decomposition $\X\approx\W\H$, where the factor matrix $\W$ contains 10\% nonzeros in the $mr$ entries and the factor matrix $\H$ contains 70\% nonzeros in the $rn$ entries.
In this dataset, we have $(m,n,r) = (100,70,3)$.
\item The vibrational signals were measured on the test rig presented in Fig.\,\ref{fig:rig}  (taken from the previous study  \cite{GABOR2023110430}).
The platform is equipped with an electric motor, gearbox, couplings, and two bearings.
One of the latter was deliberately damaged.
Diagnostic signals were measured with the accelerometer (KISTLER Model 8702B500), stacked horizontally to the shaft bearing.
The 40-second-long vibration signal was recorded with a sampling rate of 50 kHz. 
For easier visualization, one-second excerpt was selected and then transformed to a spectrogram using a 128-window length, 100-sample overlapping, and 512 frequency bins.
The raw observed signal and its spectrogram are shown in Fig.\,\ref{fig:real_signals}. 
Assuming $r = 4$ (four main components of the mixed signal), we have $(m,n,r) = (257, 1782, 4)$.
\end{itemize}

\subsection{Experimental setup}
\begin{figure}
    \centering
    \includegraphics[width=0.99\linewidth]{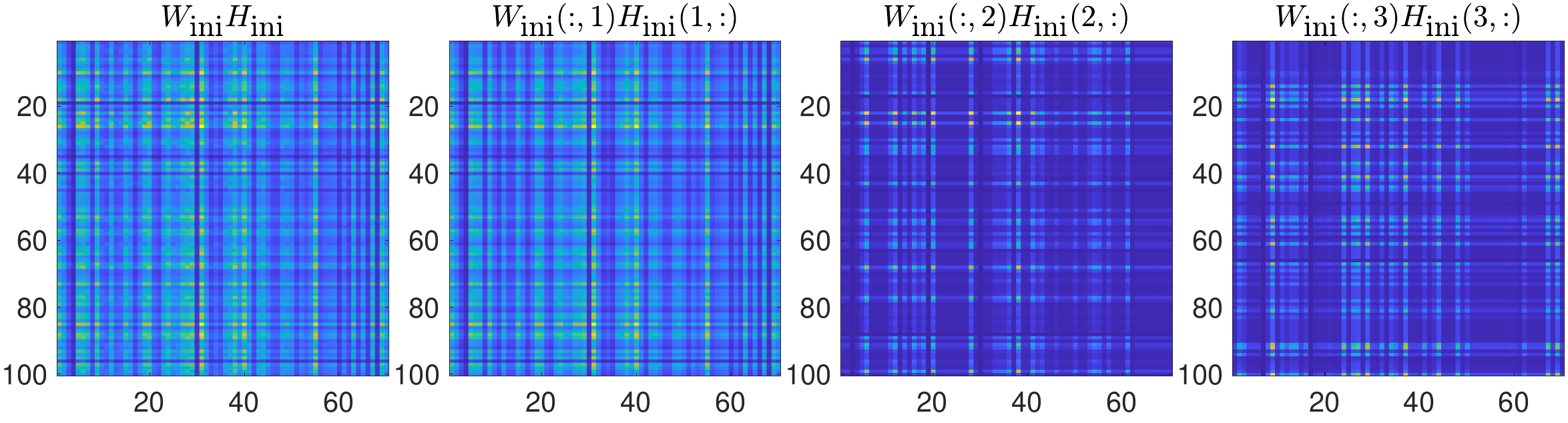}
    \caption{Initial factors generated using $(1.5|\cN|+0.5)/2$.
    Left: $\W_0\H_0$.
    Right: each rank-1 component in $\W_0\H_0$.
    }
    \label{fig:iniWH}
\end{figure}

\paragraph{\textbf{Initialization}}
For synthetic dataset, to ensure a fair comparison, all algorithms were initialized using the same initial factor matrices, which were generated from an unpenalized MU-based NMF warm start, that is initialized via nonnegative double singular value decomposition (NNDSVD) of the matrix $\X$ \cite{boutsidis2008svd}.
For real-dataset, we initialize the initial factor matrices as follows.
Let $|\cN|$ denote the truncated Gaussian distribution (i.e., each entries of the matrix follows $\cN(0,1)$ and then perform a clipping to replace negative values to 0), we generate both $\W_0, \H_0 $ using  $(1.5|\cN|+0.5)/2$.
See Fig.\,\ref{fig:iniWH}.

\paragraph{\textbf{Simulation}}
We perform 100 Monte Carlo simulations, where each run uses a different random data matrix. 
At the start of each run, the initial value of $\blambda^0$ was selected randomly following a uniform distribution $\cU[0,1]$.

\paragraph{\textbf{Normalization}}
We perform a normalization step on each column of $\W$, $\w_k$ for all $k=1, \dots, r$:
\[
\w_k = \w_k/\max(\w_k); 
\qquad    \underline{\h}_k = \underline{\h}_k*\max(\w_k).
\]

\paragraph{\textbf{Termination}}
All the algorithms were allowed to run a maximum number of iterations of $500$ for the synthetic dataset and $100$ for the real one, with an early termination tolerance of $10^{-6}$ on the relative fitting error using the IS-divergence $D_0$, defined as 
\[
\big|
D_0(\X, \W^{k+1}\H^{k+1})- D_0(\X, \W^k\H^k)
\big|
\,\big\slash\,
\big|D_0(\X, \W^k\H^k)\big|
\,\leq\,
10^{-6},
\]
where $k$ indicates the iteration of the outer loop of the algorithm. 
For the inner loop of the bi-level approach on iteration counter $t$, we stop at the maximum number of (inner) $T = 4$ iterations.
We remark that Problem \eqref{probl} is nonconvex, thus we do not have a theoretical guarantee that the algorithm's convergence can avoid poor local minima.
We only empirically show that the local minima reached are stable across different random datasets.

\paragraph{\textbf{Evaluation}}
We evaluate the algorithms according to different criteria on synthetic and real-world datasets. For both datasets, we plot the convergence of the algorithms by looking at the behavior of the Response function.

For the synthetic dataset, we evaluate the quality of the factorization with respect to the identification problem using the Signal-to-Interference
Ratio (SIR) measure \cite{cichocki2009amari} on both matrices\footnote{This measure, computed between the estimated signals and the true signals, is a log-scale similarity measure (expressed in decibels), often used in signal processing applications, and its higher value indicates a higher similarity level.}.
To highlight the effectiveness of the proposed method, we conduct a statistical comparison of SIR values across algorithms using the Kruskal-Wallis test, followed by a post-hoc multiple comparison based on the Mann-Whitney test \cite{lehmann1986testing} with Benjamini-Hochberg (BH) correction \cite{benjamini1995controlling}, maintaining a significance level of $\alpha = 0.05$. 
We also investigate the sparsity of the results obtained for the synthetic dataset using the following measure of sparsity for a generic matrix $\A\in\IRmn$:
\[
Sp(\A)\coloneqq \big(1-\|\A-10^{-6}\E\|_0 /mn\big) 100\%,
\]
where $\|\cdot\|_0$ denotes the number of non-zeros elements. 
To demonstrate the advantages of the proposed method, we analyze sparsity values across algorithms using the same statistical test as for the evaluation of the SIR (Kruskal-Wallis test, post-hoc Mann-Whitney test with BH and $\alpha = 0.05$).

On the real-world dataset, we check the goodness of the factorization by quantifying the impulsive and cyclic behavior of the signal under analysis, using a modified envelope spectrum-based indicator (ENVSI) \cite{gabor2023bearing, berrouche2024local} on time profiles. 
ENVSI can be expressed as a spectrum-based indicator (SBI):
\[
SBI \coloneqq \sum_{i=1}^{M_1}{\text{AIS}_i^2} \Bigg/ \sum_{k=1}^{M_2}{\text{S}_k^2},   
\]
where
    $\text{AIS}_i$ is the magnitude of the $i$th harmonic of the estimated signal in the frequency domain,
    $\text{S}_k$ is the magnitude in the $k$th frequency bin in the spectrum of the time profile,
    $M_1$ is the number of harmonics to be analyzed (assuming a periodic signal), and
    $M_2$ is the number of frequency bins to calculate the total energy.
SBI is zero if there are no impulsive components in the time profile, whereas a larger SBI occurs when the impulses in the time profile are stronger (which corresponds to the amplitudes in the spectrum), and the noise is weaker.
In the experiments, the number of harmonics $M_1$ is set to 6, and it was experimentally found to be sufficient for demonstrating the impulsive and periodic behavior of the SOI representing the time profile in the analyzed application. 
To demonstrate the superiority of the proposed method, we statistically compare ENVSI values across algorithms using a Kruskal-Wallis test and a multiple comparison based on the Mann-Whitney test with BH correction,  with a statistical significance level $\alpha= 0.05$.

\paragraph{\textbf{Computer}}
All the experiments were conducted in MATLAB 2021a and executed on a machine with an i7 octa-core processor and 16GB of RAM.

\subsection{Results on synthetic dataset}
\begin{figure}[ht!]
\centering
\includegraphics[width=0.65\textwidth]{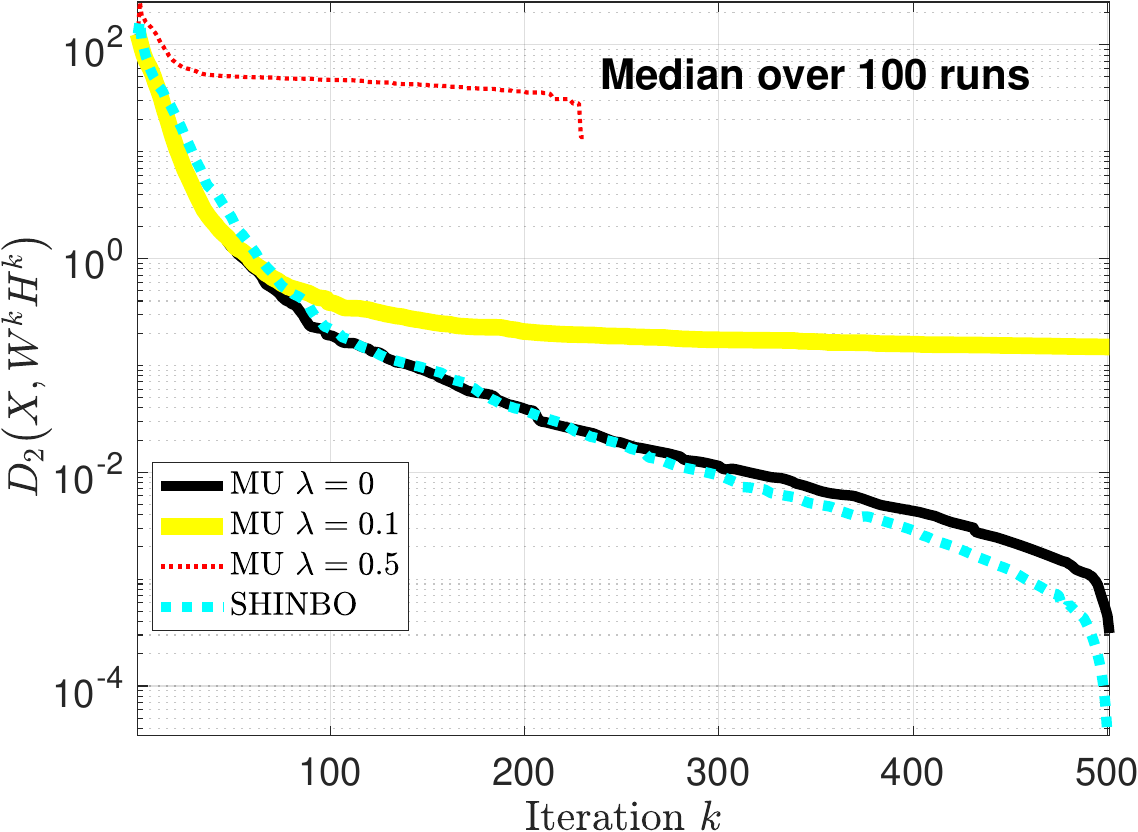}
\caption{Behavior of the Response function (outer problem) for the synthetic dataset.}
\label{fig:figures_response_A}
\end{figure}

Referring to Fig.\ref{fig:figures_response_A}, the proposed algorithm demonstrates superior performance in terms of the convergence rate of Response function compared to the other methods.
We remark that, despite SHINBO exhibiting similar, rapid convergence to unpenalized MU for the first 450 iterations (on average), the resulting decompositions have different SIR values. 
Referring to Fig.\ref{fig:figures_response_A}, we observe that SHINBO, by operating on the vector $\blambda$ across its components, achieves the highest SIR values on both matrices, confirming its effectiveness in the identification problem. At the same time, the results show higher sparsity on $\H$ and lower sparsity on 
$\W$, indicating that the automatically selected hyperparameter behaves as expected by enforcing sparsity primarily on the rows of $\H$.
On average, SHINBO outperforms the comparison algorithms with about 10\% higher SIR and 5\% higher sparsity on $\H$.

\begin{figure}[ht!]
\centering
\includegraphics[width=0.85\textwidth]{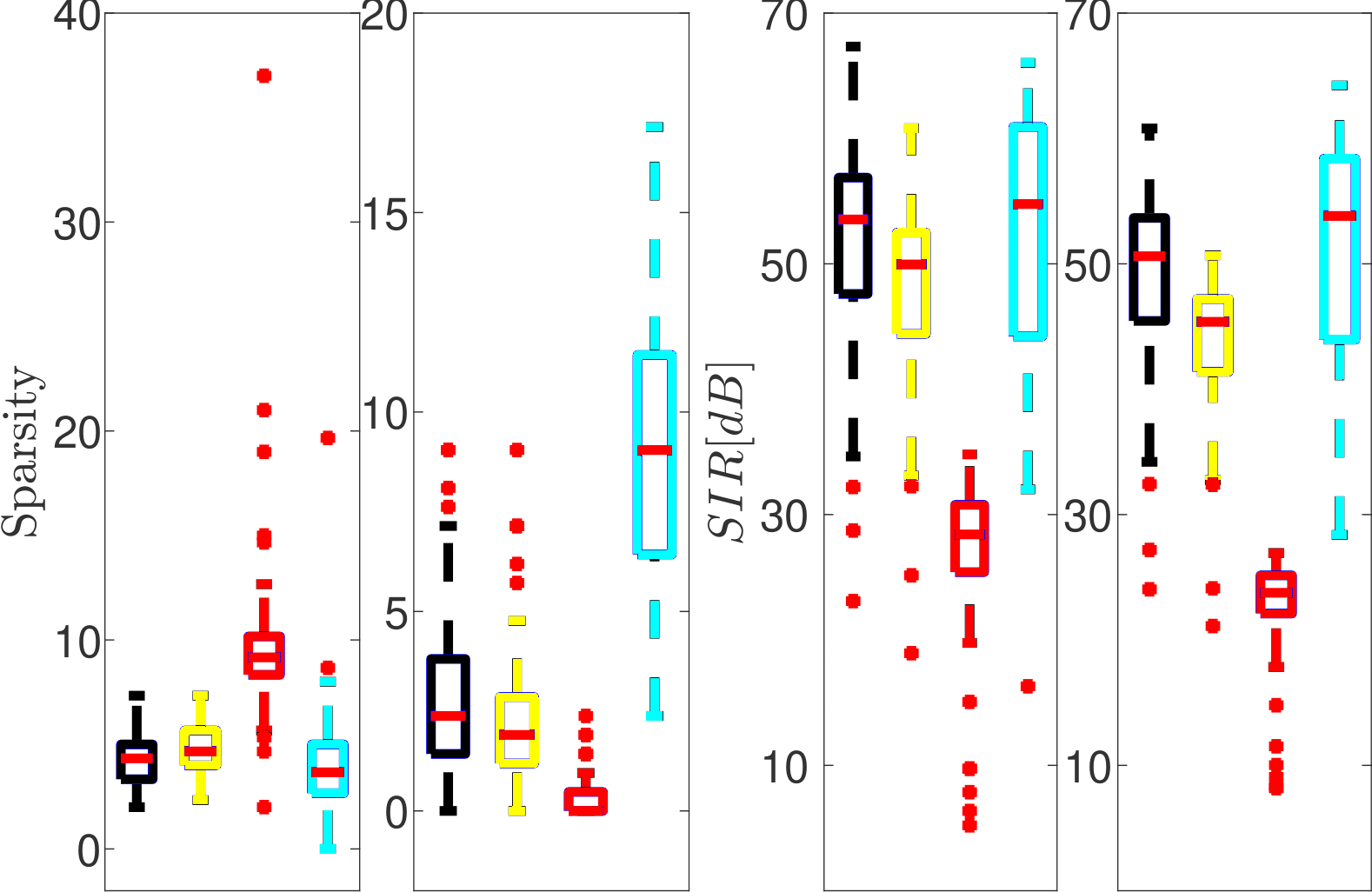}
\caption{Results of sparsity and SIR on the factor matrices on the synthetic dataset.
The left two columns are the sparsity of $\W$ (left) and $\H$ (right) of the four algorithms: from left to right are MU with $\lambda=0$, MU with $\lambda=0.1$, MU with $\lambda=0.5$, and SHINBO.
The right two columns are SIR of $\W$ (left) and $\H$ (right) of the four algorithms: from the left to right are MU with $\lambda=0$, MU with $\lambda=0.1$, MU with $\lambda=0.5$, and SHINBO.
}
\label{fig:SIR_SP}
\end{figure}

These results are also confirmed by the statistical comparisons. 
The Kruskal-Wallis tests present $p$-values lesser than $10^{-14}$ and the details of the pairwise comparison with a BH comparison are presented in the tables \ref{tab:sir_A} and \ref{tab:sparsity_A}.
Table \ref{tab:sir_A} reports the results of pairwise comparisons for the SIR coefficients on 
($\W$, $\H$). The significantly low 
$p$-values indicate that SHINBO consistently achieves higher SIR values compared to MU under different penalization settings, thereby confirming its superior accuracy in handling the identification problem in matrix reconstruction.
Similarly, Table \ref{tab:sparsity_A} presents the $p$-values obtained from pairwise comparisons of sparsity coefficients on ($\W$, $\H$). 
The results show that SHINBO exhibits statistically significant differences with respect to MU variants, highlighting its ability to balance accuracy and sparsity in both synthetic and real datasets.

\begin{table*}[h!]\caption{$p$-values of pairwise comparisons for SIR coefficients on ($\W, \,\H$). 
In the table $\epsilon = 10^{-16}$.}
\label{tab:sir_A}
\centering
\begin{tabular}{c| c c c }
& MU & MU $\lambda=0.1$ & MU $\lambda=0.5$\\ \hline
MU $\lambda=0.1$ & $8.1 \cdot 10^{-5}$, $6.3 \cdot 10^{-10}$ & -  & - \\
MU $\lambda=0.5$ & $<2\epsilon$,$<2\epsilon$ & $<2\epsilon$,$<2\epsilon$ & - \\
SHINBO & 0.14,0.0042 & $5.4 \cdot 10^{-5},1.2  \cdot 10^{-9}$ & $<2\epsilon$, $<2\epsilon$ 
\end{tabular}
\end{table*}
 
\begin{table*}[h!]\caption{$p$-values of pairwise comparisons for sparsity coefficients on $(\W,\H)$.}
        \label{tab:sparsity_A}
            \centering
            \begin{tabular}{c| c c c }
                 & MU & MU $\lambda=0.1$ & MU $\lambda=0.5$\\
                 \hline
            MU $\lambda=0.1$ & 0.00036, 0.0075 & -  & - \\
            MU $\lambda=0.5$ & $<2\epsilon$, $<2\epsilon$ & $<2\epsilon$,$<2\epsilon$ & - \\
            SHINBO  & 0.023, $<2\epsilon$  & $3.12 \cdot 10^{-6}$, $<2\epsilon$  & $<2\epsilon, <2\epsilon$
            \end{tabular}    
	\end{table*}

\paragraph{\textbf{Sensitivity to the rank}}
We performed a sensitivity analysis with respect to the rank $r$ for the synthetic dataset.
Tables \ref{tab:rH}-\ref{tab:rW} show that across different values of $r$, the SHINBO algorithm performs better on SIR average than the other compared methods.
We note that SHINBO also has the highest variance among all tested methods.
Finally, we observed that the mean$\pm$std values of SIR decreases as rank $r$ increases for all method.
All these observations can be explained as follows.
\begin{itemize}
\item First, MU $\lambda=0.5$ has the smallest variance across all methods, but also at the same time has the lowest SIR on average, meaning the method has the worst performance.
\item The low variance of MU $\lambda=0.5$ is contributed to the fact that the method has a relatively larger $\lambda$ compared to the baseline MU method.
Recall that the diversity measure \eqref{J_row}, such as $J(\A) = \sum_{i=1}^n \|\A_{i:}\|_1^2$, is applied on $\H$, thus a higher $\lambda$ in MU will made the method focusing more on minimizing the column norm of $\H$.
A solution with small column-wise norm will naturally has smaller variance, and thus this explains why MU $\lambda = 0.5$ has the smallest variance.
In general, if $\lambda$ is huge (e.g. $\lambda=10,100,...$), then we can expect that MU with such value will have even smaller variance on the SIR results.
As Problem\,\eqref{probl} contains two terms, thus a large $\lambda$ will lead to the MU focusing on column norm minimization instead of the factorization, leading to a lower SIR in reconstruction.
\item Among the MU methods, MU with $\lambda=0$ has the highest SIR on average and also the highest variance.
Compared with MU, SHINBO has a higher SIR on average in this case.
\item Note that SHINBO needs to find more components of $\blambda=(\lambda_1,\lambda_2,...)$, instead of just one $\lambda$ for MU.
For example, for $r=3$, SHINBO produces a value $\blambda=(0.0142,0.0292,0.0305)$.
With an increasing $r$, the number of components of $\blambda$ also increase, making Problem\,\eqref{probl} harder to solve for SHINBO.
Under such situation, SHINBO is still able to have on average higher SIR.
\end{itemize}

\begin{table*}[h!]\caption{Mean$\pm$std of SIR values of $\H$ for the synthetic dataset, w.r.t. values of $r$.}
        \label{tab:rH}
            \centering
            \begin{tabular}{c| c c c c}
              rank   & MU & MU $\lambda=0.1$ & MU $\lambda=0.5$ & SHINBO \\
                 \hline
            4 & $39.096 \hspace{-1mm}\pm\hspace{-1mm} 14.023$
  & $35.899 \hspace{-1mm}\pm\hspace{-1mm} 11.641$
  & $17.457 \hspace{-1mm}\pm\hspace{-1mm} 7.164$
  & $44.734 \hspace{-1mm}\pm\hspace{-1mm} 17.894$ \\
            5 &  $29.830 \hspace{-1mm}\pm\hspace{-1mm} 12.216$
  & $28.183 \hspace{-1mm}\pm\hspace{-1mm} 11.050$
  & $14.769 \hspace{-1mm}\pm\hspace{-1mm} 6.545$
  & $35.801 \hspace{-1mm}\pm\hspace{-1mm} 16.903$\\
            6 &  $23.423 \hspace{-1mm}\pm\hspace{-1mm} 9.159$
  & $22.576 \hspace{-1mm}\pm\hspace{-1mm} 8.694$
  & $11.849 \hspace{-1mm}\pm\hspace{-1mm} 5.028$
  & $26.930 \hspace{-1mm}\pm\hspace{-1mm} 13.764$ 
            \end{tabular}    
	\end{table*}

    \begin{table*}[h!]\caption{Mean$\pm$std of SIR values of $\W$ for the synthetic dataset, w.r.t. values of $r$.}
        \label{tab:rW}
            \centering
            \begin{tabular}{c| c c c c}
              rank   & MU & MU $\lambda=0.1$ & MU $\lambda=0.5$ & SHINBO \\
                 \hline
            4 & $41.836 \hspace{-1mm}\pm\hspace{-1mm} 17.689$
  & $39.183 \hspace{-1mm}\pm\hspace{-1mm} 15.542$
  & $20.736 \hspace{-1mm}\pm\hspace{-1mm} 10.644$
  & $46.108 \hspace{-1mm}\pm\hspace{-1mm} 20.023$ \\
            5 &  $30.665 \hspace{-1mm}\pm\hspace{-1mm} 15.257$
  & $29.489 \hspace{-1mm}\pm\hspace{-1mm} 13.976$
  & $17.347 \hspace{-1mm}\pm\hspace{-1mm} 9.014$
  & $36.755 \hspace{-1mm}\pm\hspace{-1mm} 18.479$\\
            6 &  $23.879 \hspace{-1mm}\pm\hspace{-1mm} 10.719$
  & $23.317 \hspace{-1mm}\pm\hspace{-1mm} 10.242$
  & $14.153 \hspace{-1mm}\pm\hspace{-1mm} 6.501$
  & $27.983 \hspace{-1mm}\pm\hspace{-1mm} 15.461$ 
            \end{tabular}    
	\end{table*}

{\paragraph{\textbf{Robustness under measurement noise}}
To assess robustness to measurement noise, we corrupted the clean data $\X$ by adding i.i.d. zero-mean Gaussian noise and then projecting onto the nonnegative orthant. Formally, we generated
\[
\Y = \max(\X + \varepsilon\mathcal{N}(0,1),0),
\]
where $\varepsilon>0$ denotes the noise standard deviation. 
The SIR metrics on both factors in Tables~\ref{tab:sir_noise}-\ref{tab:sir_noise2} show a gradual degradation of SIR as $\varepsilon$ increases (i.e., as the noise level grows), without abrupt performance breakdowns. Across all tested noise conditions, SHINBO remains comparable to the MU baselines, often slightly superior at a low noise level (e.g., $\varepsilon=0.01$), and on par at higher noise, thus substantiating its robustness to measurement noise.

\begin{table*}[h!]\caption{Mean$\pm$std of SIR values for $\H$, with respect to different noise.}
         \label{tab:sir_noise}
             \centering
             \begin{tabular}{c| c c c c}
               $\varepsilon$   & MU & MU $\lambda=0.1$ & MU $\lambda=0.5$ & SHINBO \\
                  \hline
            0.01 &  $27.3259 \hspace{-1mm}\pm\hspace{-1mm} 4.7506$
   & $27.3571 \hspace{-1mm}\pm\hspace{-1mm} 4.5884$
   & $20.9436 \hspace{-1mm}\pm\hspace{-1mm} 4.4491$
   & $27.6639 \hspace{-1mm}\pm\hspace{-1mm} 5.1397$ \\
             0.05 &  $19.5527 \hspace{-1mm}\pm\hspace{-1mm} 3.7249$
   & $19.5402 \hspace{-1mm}\pm\hspace{-1mm} 3.7146$
   & $16.3236 \hspace{-1mm}\pm\hspace{-1mm} 4.3636$
   & $19.2795 \hspace{-1mm}\pm\hspace{-1mm} 4.1752$ \\
             0.1 &  $15.6428 \hspace{-1mm}\pm\hspace{-1mm} 3.3477$
   & $15.6939 \hspace{-1mm}\pm\hspace{-1mm} 3.2599$
   & $12.6351 \hspace{-1mm}\pm\hspace{-1mm} 4.5464$
   & $15.3858 \hspace{-1mm}\pm\hspace{-1mm} 3.3559$ 
 
            \end{tabular}    
 	\end{table*}

    \begin{table*}[h!]\caption{{Mean$\pm$std of SIR values for $\W$, with respect to different noise.}}
         \label{tab:sir_noise2}
             \centering
              \begin{tabular}{c| c c c c}
               $\varepsilon$   & MU & MU $\lambda=0.1$ & MU $\lambda=0.5$ & SHINBO \\
               \hline
            0.01 &  $28.3267 \hspace{-1mm}\pm\hspace{-1mm} 4.7552$
   & $28.4009 \hspace{-1mm}\pm\hspace{-1mm} 4.7298$
   & $23.8437 \hspace{-1mm}\pm\hspace{-1mm} 6.1478$
   & $29.0103 \hspace{-1mm}\pm\hspace{-1mm} 4.8775$ \\
             0.05 &  $19.8840 \hspace{-1mm}\pm\hspace{-1mm} 3.6120$
   & $19.8576 \hspace{-1mm}\pm\hspace{-1mm} 3.6338$
   & $17.3408 \hspace{-1mm}\pm\hspace{-1mm} 5.4629$
   & $19.3763 \hspace{-1mm}\pm\hspace{-1mm} 4.1957$ \\
             0.1 &  $15.7853 \hspace{-1mm}\pm\hspace{-1mm} 3.1266$
   & $15.8039 \hspace{-1mm}\pm\hspace{-1mm} 3.1181$
   & $13.2717 \hspace{-1mm}\pm\hspace{-1mm} 5.1220$
   & $15.6706 \hspace{-1mm}\pm\hspace{-1mm} 3.1455$ 
 
            \end{tabular}  
 	\end{table*}

\subsection{Results on real-world dataset}

We evaluate the proposed IS-NMF model with row-wise penalty adaptation with SHINBO on a vibration signal from a faulty rolling bearing (Fig.~\ref{fig:real_signals}). Following the experimental setup, NMF is applied to the signal spectrogram, and the temporal activations (rows of $\H$) are shown in Fig.~\ref{fig:profiles}(a).
One activation (here, the second component) exhibits a sparse, periodic, spike-like pattern characteristic of the SOI, i.e., fault-induced impulsivity in the bearing. Because of NMF’s permutation ambiguity, the SOI can appear in any row; the remaining rows capture latent components associated with broadband and transient disturbances.

Fig.~\ref{fig:profiles}(b) displays the envelope spectra of the activations from Fig.~\ref{fig:profiles}(a). The first harmonic at approximately $91$Hz aligns with the shaft rotational speed, corroborating that the second component corresponds to the SOI. To quantify spectral leakage and identifiability, the ENVSI metric was used.  Fig.~\ref{fig:envsi1} summarizes the quantitative results. The right panel shows that SHINBO attains the highest ENVSI, with values above $0.77$ and fewer outliers, indicating reliable SOI recovery consistent with our prior knowledge about the physical state of the tested rolling bearing. The left panel reports the convergence of the Response function, where the proposed method outperforms the baselines.

\begin{figure}[ht!]
\centering
\includegraphics[width=0.99\textwidth]{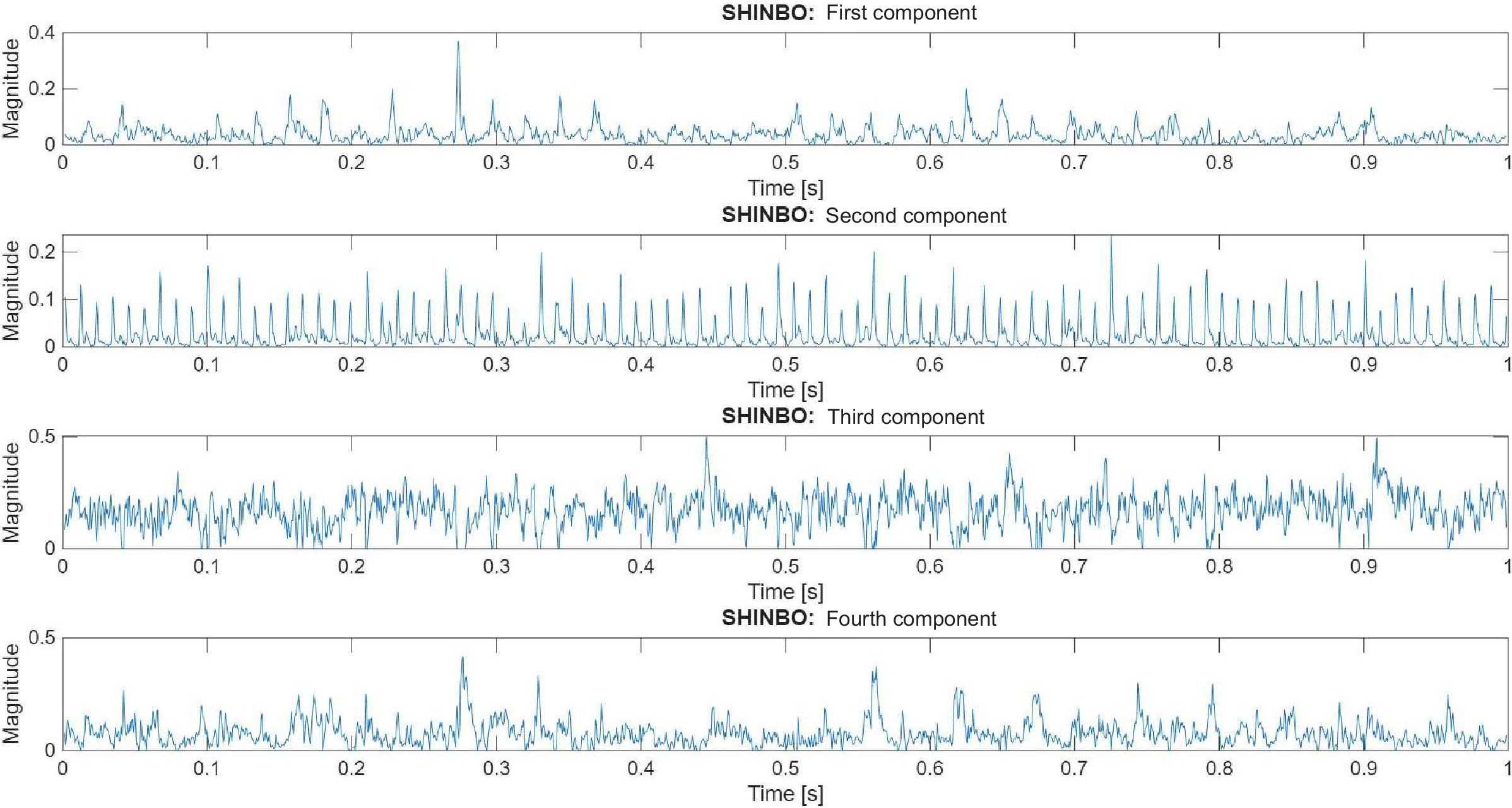} \\
\hspace{0.1cm} (a) \\
\includegraphics[width=0.99\textwidth]{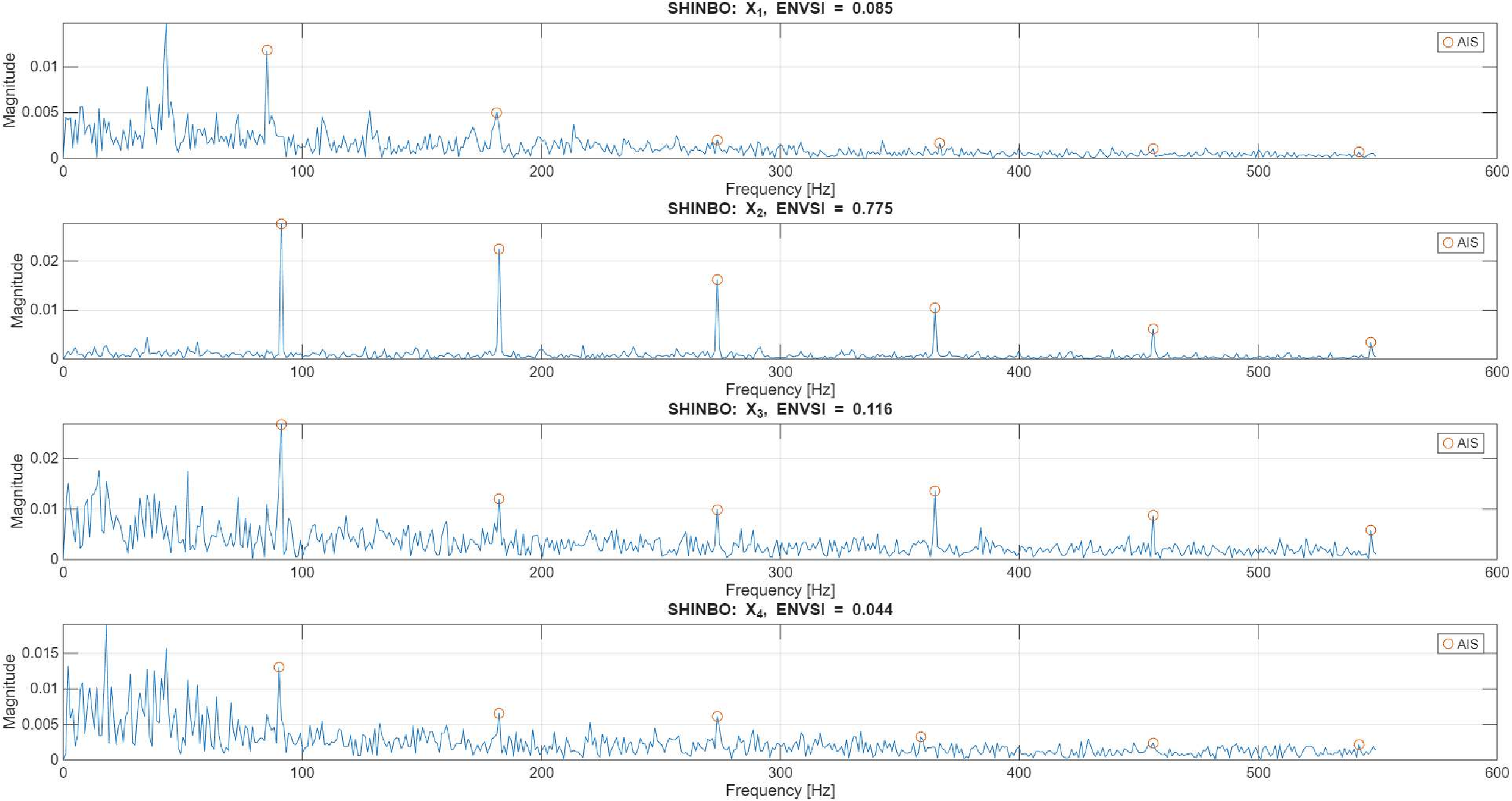}
\hspace{5cm} (b) \\
\caption{(a) Time-domain components (the rows of $\H$) estimated with SHINBO from the faulty signal. 
(b) Envelope spectrum of the estimated time-domain components. The respective ENVSI scores are given in the subtitles. }
\label{fig:profiles}
\end{figure}

\begin{figure}[ht!]
\centering
\includegraphics[width=0.7\textwidth]{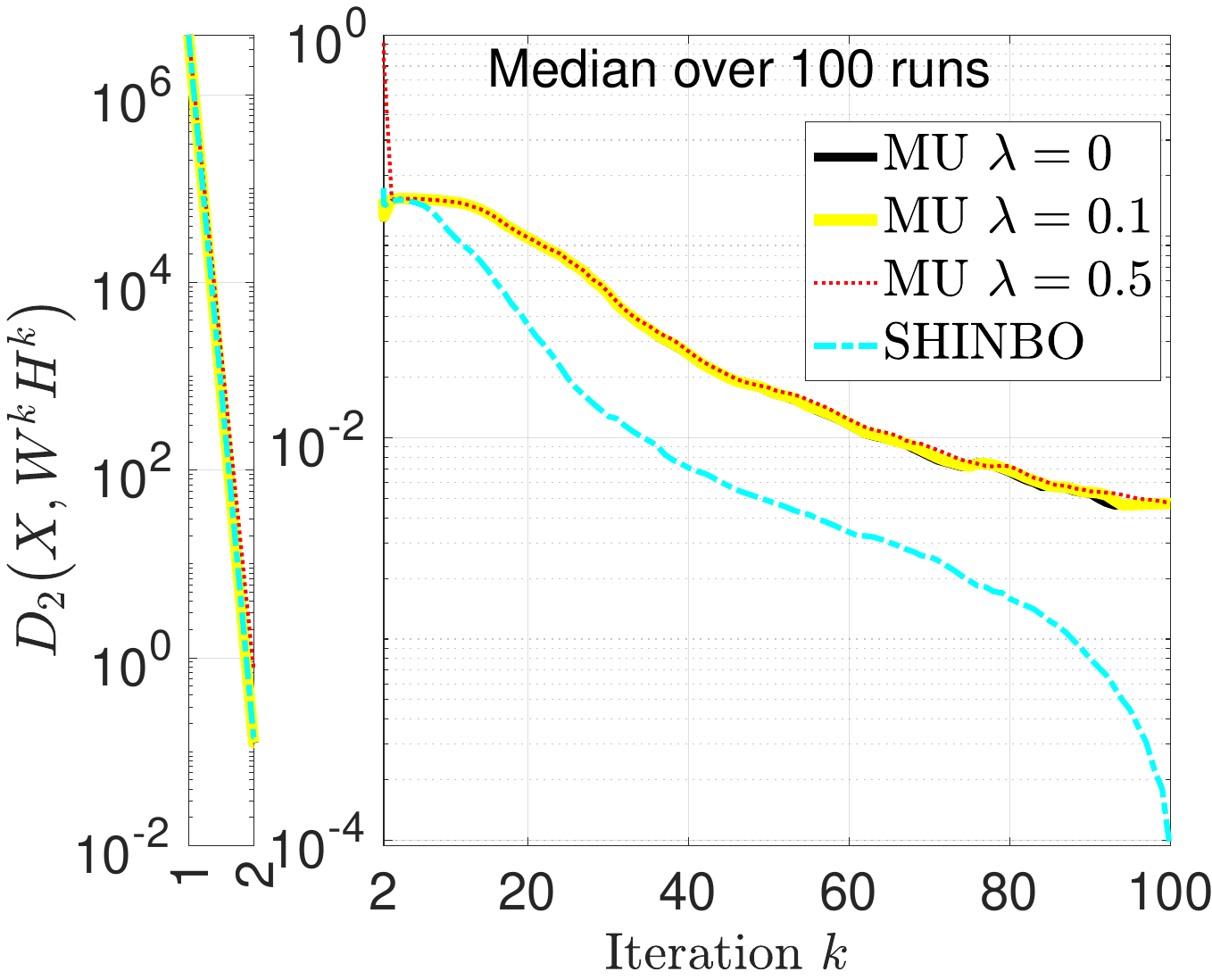}
\includegraphics[width=0.235\textwidth]{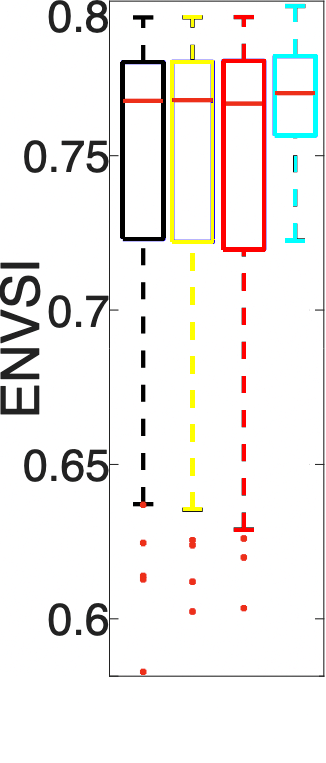}
\caption{
Left: Behavior of the Response function (outer problem) for the real dataset.
Right: The ENVSI score of $\H$ of the four algorithms: from left to right are MU with $\lambda=0$, MU with $\lambda=0.1$, MU with $\lambda=0.5$, and SHINBO.
}
\label{fig:envsi1}
\end{figure}

These results are also confirmed by the statistical comparisons.
The Kruskal-Wallis tests present $p$-values lesser than the fixed significant level (0.04), and the details of the pairwise comparison with a BH comparison are presented in Table \ref{tab:envsi}, which summarizes the results for the ENVSI coefficient. 
The $p$-values indicate no significant differences among the MU variants, whereas SHINBO shows statistically significant improvements with respect to all MU configurations, further confirming its effectiveness in capturing relevant structural information.

\begin{table*}[h!]\caption{$p$-values results of pairwise comparisons for ENVSI coefficients.}
        \label{tab:envsi}
        \centering
            \begin{tabular}{c| c c c }
                 & MU & MU $\lambda=0.1$ & MU $\lambda=0.5$\\
                 \hline
            MU $\lambda=0.1$ & 0.98 & -  & - \\
            MU $\lambda=0.5$ & 0.98 & 0.98  & - \\
            SHINBO  & 0.04 & 0.04 & 0.04 
            \end{tabular}
	\end{table*}

The concept is that SHINBO is designed to effectively identify which components of $\blambda$ are associated with noise and which are linked to the true sparse signal.
The goal is to apply greater penalization to certain components, thereby filtering out the noise while preserving the component representing the SOI. The results indicate that SHINBO successfully isolates the meaningful signal while suppressing irrelevant noise. 



\section{Conclusion}\label{sec:conl}
In this work, we addressed the critical challenge of selecting penalty hyperparameters in NMF by introducing SHINBO, a novel algorithm that employs a bi-level optimization framework to adaptively tune row-dependent penalties. 
The core idea is that SHINBO does not require setting the penalty coefficient value as a fixed parameter in advance; instead, it automatically optimizes the hyperparameter during the process. Compared to standard algorithms, where the penalty coefficient value is user-dependent, SHINBO consistently achieves high SIR, sparsity, and ENVSI on synthetic and real datasets in a fully automatic manner. This demonstrates that the proposed method can solve the problem in a one-shot way, without relying on any prior parameter tuning, while still providing optimal results in terms of: i)identification: SHINBO yields optimal SIR for both reconstructed matrices; ii) Sparsity: it respects the constraints imposed by the penalizations on the matrix $\mathbf{H}$; iii)Spectral analysis: it ensures an effective analysis of the signal spectrum.
By focusing on the IS divergence, SHINBO proves highly effective for extracting low spectral density components in spectrograms, particularly in the presence of noise. 
The ability of the algorithm to enforce sparsity constraints and dynamically adjust penalties ensures a more precise separation of meaningful signals from noisy disturbances.  

Through experiments on both synthetic and real-world datasets, SHINBO demonstrated its superior performance compared to traditional NMF methods.
For real-world applications, such as noninvasive vibration-based fault detection in rolling bearings, SHINBO excelled at isolating sparse, periodic signal components in high-frequency subbands, even when heavily masked by broader noise.

Overall, the results highlight SHINBO's potential to significantly improve signal recovery in complex, noise-dominated environments.
By tackling the hyperparameter selection problem with an adaptive, data-driven approach, SHINBO not only advances the field of NMF but also provides a robust tool for applications requiring precise spectral decomposition and noise suppression.
Future work will explore the scalability of SHINBO for larger datasets and its adaptability to other domains with similar challenges.

\section*{Acknowledgment}
The authors thank Professor Radoslaw Zimroz from Faculty of Geoengineering, Mining and Geology at Wroclaw University of Science and Technology, for providing the real data collected in his laboratory. 

N.D.B., F.E., L.S. are members of the Gruppo Nazionale Calcolo Scientifico - Istituto Nazionale di Alta Matematica (GNCS-INdAM). 

\section*{Funding}
N.D.B., F.E., and L.S. are partially supported by ``INdAM - GNCS Project'', CUP: E53C24001950001. 
N.D.B. and F.E. are supported by Piano Nazionale di Ripresa e Resilienza (PNRR), Missione 4 ``Istruzione e Ricerca''-Componente C2 Investimento 1.1, ``Fondo per il Programma Nazionale di Ricerca e Progetti di Rilevante Interesse Nazionale'', Progetto PRIN-2022 PNRR, P2022BLN38, Computational approaches for the integration of multi-omics data. CUP: H53D23008870001.

\section*{Authors contribution}
All authors contributed equally to this work.

\section*{Conflict of interest}
The authors have no relevant financial interest to disclose.

\section*{Declaration of generative AI and AI-assisted technologies}
The authors used ChatGPT 5.1 (OpenAI) in order to revise the English grammar. After using this tool/service, the authors reviewed and edited the content as needed and take full responsibility for the content of the published article.

\bibliographystyle{elsarticle-num}
\bibliography{reference}
\end{document}